\documentclass[runningheads]{llncs}
\usepackage{graphicx}
\usepackage[font=small,skip=1pt]{caption}
\usepackage{amsmath,amssymb} 
\usepackage{color}
\usepackage[width=122mm,left=12mm,paperwidth=146mm,height=193mm,top=12mm,paperheight=217mm]{geometry}

\usepackage{times}
\usepackage{epsfig}
\usepackage{bigstrut}
\usepackage{tablefootnote}
\usepackage{multirow}
\usepackage{algorithm}
\usepackage{algorithmic}
\usepackage{subfigure}
\usepackage{pifont}
\newcommand{\cmark}{\ding{51}}%
\newcommand{\xmark}{\ding{55}}%
\usepackage[pagebackref=true,breaklinks=true,letterpaper=true,colorlinks,bookmarks=false]{hyperref}

\usepackage{xspace}
\makeatletter
\DeclareRobustCommand\onedot{\futurelet\@let@token\@onedot}
\def\@onedot{\ifx\@let@token.\else.\null\fi\xspace}
\def\eg{\emph{e.g}\onedot} 
\def\ie{\emph{i.e}\onedot} 
 
\def\etc{\emph{etc}\onedot} 
 
\def\etal{\emph{et al}\onedot}
\makeatother

\newcommand*\samethanks[1][\value{footnote}]{\footnotemark[#1]}

\usepackage{array}
\newcolumntype{C}[1]{>{\centering\let\newline\\\arraybackslash\hspace{0pt}}m{#1}}

\begin{document}
\pagestyle{headings}
\mainmatter
\def\ECCV18SubNumber{1423}  

\title{Person Search via A Mask-Guided Two-Stream CNN Model} 

\titlerunning{Person Search via A Mask-guided Two-stream CNN Model}

\authorrunning{D. Chen \etal}

\author{Di Chen\inst{1} \and Shanshan Zhang\inst{1}\thanks{Corresponding authors.} \and Wanli Ouyang\inst{2} \and Jian Yang\inst{1}\samethanks \and Ying Tai\inst{3}}

\institute{
PCA Lab, Key Lab of Intelligent Perception and Systems for High-Dimensional Information of
Ministry of Education, and Jiangsu Key Lab of Image and Video Understanding \\ for Social Security, \\
School of Computer Science and Engineering,
Nanjing University of Science and Technology\\
 \and 
The University of Sydney, Sensetime Intelligent Multimedia Laboratory Group
\\
 \and 
Youtu Lab,
Tencent\\
\email{\{dichen, shanshan.zhang, csjyang\}@njust.edu.cn, wanli.ouyang@sydney.edu.au, yingtai@tencent.com}
}

\maketitle

\begin{abstract}
In this work, we tackle the problem of person search, which is a challenging task consisted of pedestrian detection and person re-identification~(re-ID). Instead of sharing representations in a single joint model, we find that separating detector and re-ID feature extraction yields better performance.
In order to extract more representative features for each identity, 
we propose a simple yet effective re-ID method, which models foreground person and original image patches individually, and obtains enriched representations from two separate CNN streams.
On the standard person search benchmark datasets, we achieve mAP of $83.0\%$ and $32.6\%$ respectively for CUHK-SYSU and PRW, surpassing the state of the art by a large margin (more than 5pp).

\keywords{Person search; Pedestrian detection; Person re-identification; Foreground}
\end{abstract}

\section{Introduction}
The task of person search is first introduced by~\cite{Xu2014}, which unifies the pedestrian detection and person re-identification in a coherent system. A typical person re-identification method aims to find matchings between the query probe and the cropped person image patches from the gallery, thus requiring perfect person detection results, which are hard to obtain in practice. In contrast, person search, which searches the queried person over the whole image instead of comparing with manually cropped person image locally, is closer to the real world applications.
However, considering the tasks of detection and re-ID together brings domain-specific difficulties: large appearance variance across cameras, low resolution, occlusion, \etc. In addition, sharing features between detection and re-ID also accumulates errors from each of them, \eg false alarms, misalignments and inexpressive person descriptors, which further jeopardizes the final person search performance.

Following~\cite{Xu2014}, a few other works~\cite{Zheng_2017_CVPR,Xiao_2017_CVPR,Xiao2017IANTI,Liu_2017_ICCV} have also been proposed for person search. Most of them~\cite{Xiao_2017_CVPR,Xiao2017IANTI,Liu_2017_ICCV} focus on an end-to-end solution based on Faster R-CNN~\cite{Ren2017}. Specifically, an axillary fully-connected (FC) layer is added upon the top convolutional layer of Faster R-CNN to extract discriminative features for re-identification. During training, they optimize a joint loss which is composed of the Faster R-CNN losses and a person categorization loss. However, we argue that it is not appropriate to share representations between the detection and re-ID tasks, as their goals contradict with each other. For the detection task, all people are treated as one class, and the goal is to distinguish them from the background, thus the representations focus on the commonness of different people, \eg the body shape; while for the re-ID task, different people are deemed as different categories, and the goal is to maximize the differences in between, thus the representations focus on the characteristics of each identity, \eg clothing, hairstyle, \etc. In other words, the detection and re-ID tasks aim to model the inter-class and intra-class difference for people respectively. Therefore, it makes more sense to separate the two tasks rather than solving them jointly.

In the community of person re-ID, it is widely believed that discriminative information lies in foreground, while background is one of the detrimental factors and ought to be neglected or removed during feature extraction~\cite{bg_removal_1,bg_removal_2}. An intuitive idea would be to extract features on the foreground person patch only while ignoring the background area. However, simply abandoning all background information may harm the re-ID performance from two aspects. Firstly, the feature extraction procedure may gather errors from imperfect or noisy segmentation masks, \ie identification information loss caused by fractional body shape. Secondly, background information sometimes acts as useful context, \eg attendant suitcases, handbags or companions. Casting out all background area would neglect some informative cues for the re-ID problem.
Therefore, we argue that it is more suitable to consider a compromised strategy of paying extra attention on the foreground person while also using the background as a complementary cue.

\begin{figure*}[t]
\begin{center}
\includegraphics[width=1.0\linewidth]{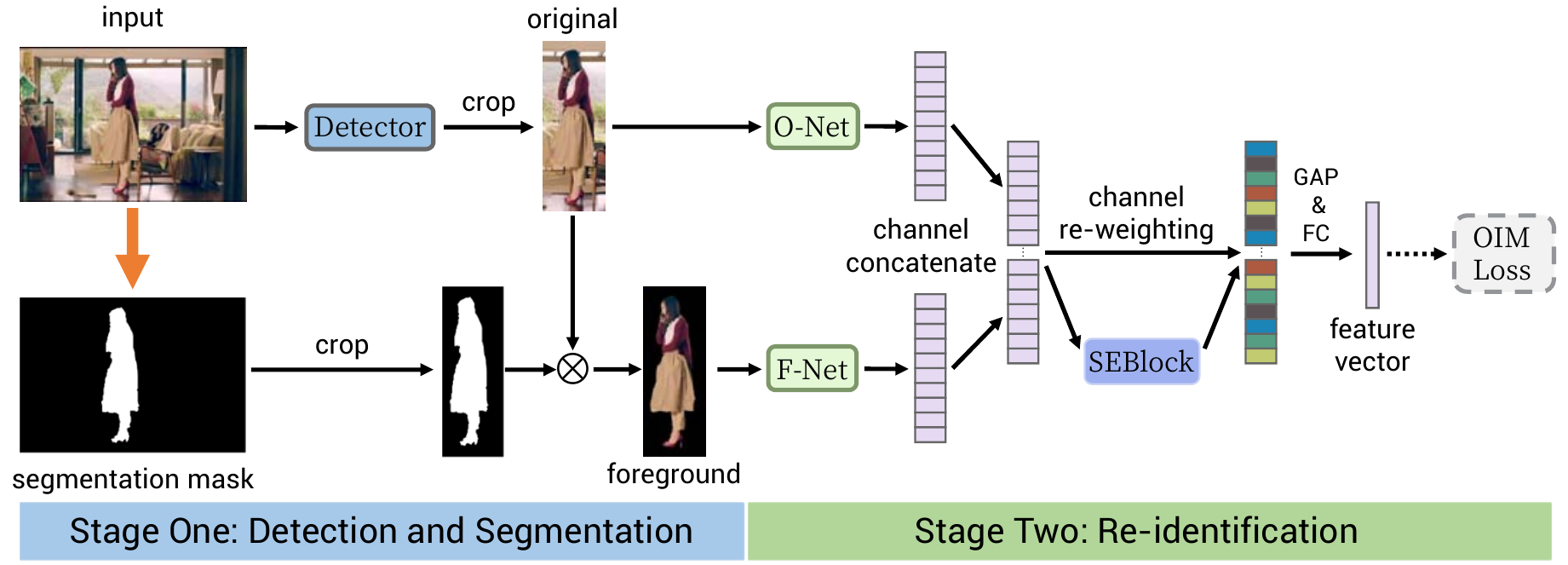}
\end{center} 
\caption{The proposed framework for person search. It is composed of two stages: 1) Detection and segmentation. We use an adapted  Faster R-CNN~\cite{CityPersons2017cvpr} as our pedestrian detector; the segmentation mask is generated by a MS COCO pre-trained FCIS model~\cite{Li_2017_CVPR} without any fine-tuning; 2) Re-identification. The feature extractor is supervised by an Online Instance Matching~(OIM) loss~\cite{Xiao_2017_CVPR}. Please note that the detector and re-ID feature extractor are trained independently.} 
 
\label{fig:pipeline}
\end{figure*}

Inspired by the above discussions, we propose a new approach for person search. It consists of two stages: pedestrian detection and person re-identification. We solve them separately, without sharing any representations. Furthermore, we propose a Two-stream CNN to model foreground person and original image independently, which aims to extract more informative features for each identity and still consider the complementarity of the background.
The whole framework is demonstrated in Fig.~\ref{fig:pipeline}, and we will talk about more details in Sec. \ref{sec:method}. 

In summary, our contributions are three-folds:
\begin{itemize}
\item[$\bullet$] To the best of our knowledge, this paper is the first work showing that for the person search problem, better performance can be achieved by solving the pedestrian detection and person re-identification tasks separately rather than jointly.
\item[$\bullet$] We propose a Mask-guided Two-Stream CNN Model~(MGTS) for person re-id, which explicitly makes use of one stream from the foreground as the emphasized information and enriches the representation by incorporating another separate stream from the original image.
\item[$\bullet$] Our proposed method achieves mAP of $83.0\%$ and $32.6\%$ on CUHK-SYSU~\cite{Xiao_2017_CVPR} and PRW~\cite{Zheng_2017_CVPR} benchmarks respectively, which improves over the previous state-of-the-arts by a large margin (more than 5pp).
\end{itemize}


\section{Related Work}
We first review existing works on person search, which is a recently proposed topic. Since our person search method is composed of two stages: pedestrian detection and person re-identification, we also review some recent works in both fields.

\noindent\textbf{Person search.} Person search has drawn much research interest since the publication of two large scale datasets: CUHK-SYSU~\cite{Xiao_2017_CVPR} and PRW~\cite{Zheng_2017_CVPR}. Zheng \etal~\cite{Zheng_2017_CVPR} conduct a detailed survey on various \textit{separated models} and propose to solve the person search problem in two separate models, one for detection and another for re-ID.
Other works propose to solve the problem in an \textit{end-to-end fashion} by employing the Faster R-CNN detector~\cite{Ren2017} for pedestrian detection and share the base network between detection and re-identification~\cite{Xiao_2017_CVPR}. In \cite{Xiao2017IANTI}, feature discriminative power is increased by introducing center loss~\cite{Wen2016} during training. Liu \etal~\cite{Liu_2017_ICCV} improve the localization policy of Faster R-CNN by recursively shrink the search area from the whole image till achieving precise location of the target person.
In this paper, we first made a systematic comparison between \textit{separated models} and \textit{joint models}, and show that a separated solution improves both the detection and re-identification results.

\noindent\textbf{Pedestrian detection.} Pedestrian detection is canonical object detection, especially when hand-crafted features are widely used. The classic HOG descriptor \cite{Dalal2005Cvpr} is based on local image differences, and successfully represents the special head-shoulder shape of pedestrians. A deformable part model (DPM) \cite{DPM} is proposed to handle deformations and still uses HOG as basic features. More recently, the integral channel feature (ICF) detectors \cite{Dollar2009,shanshan2014cvpr,ICF2015Cvpr} become popular, as they achieve remarkable improvements while running fast. In recent years, convnets are also employed in pedestrian detection and further push forward the progress \cite{shanshan_cvpr16,shanshan2018pami,Ouyang2013JointDeep,Ouyang:DBNHuman}. Some works use the R-CNN architecture, which relies on ICF for proposal generation \cite{shanshan_cvpr16,shanshan2018pami}. Aiming for an end-to-end procedure, Faster R-CNN\cite{Ren2017} is adopted and it achieves top results by applying proper adaptations \cite{CityPersons2017cvpr,shanshan2018cvpr}. Therefore, we use the adapted Faster R-CNN detector in this paper.

\noindent\textbf{Person re-ID.} Early person re-identification methods focus on manually designing discriminative features~\cite{Wang2007,conf/eccv/GrayT08,Farenzena2010,Zhao2013,Liao_2015_CVPR}, using salient regions \cite{zhao2017person}, and learning distance metrics~\cite{Kostinger2012,Li2015,Liao2015,Paisitkriangkrai2015,Zhang_2016_CVPR}. For instance, Zhao \etal~\cite{Zhao2013} propose to densely combine color histogram and SIFT features as the final multi-dimensional descriptor vector. Kostinger \etal~\cite{Kostinger2012} present KISS method to learn a distance metric from equivalence constraints. CNN-based models have attracted extensive attentions since the successful applications by two pioneer works~\cite{Yi2014,Li2014}. Most of those CNN models can be categorized into two groups. The first group uses the siamese model with image pairs~\cite{Yi2014,Li2014,Ahmed2015,Varior2016,Liu2017,xu2018attention} or triplets~\cite{Ding2015,Cheng2016} as inputs. The main idea of these works is to minimize the feature distance between the same person and maximize the distance between different people. The second group of works formulate the re-identification task as a classification problem~\cite{xiao2016learning,Zheng2016,Zheng_2017_CVPR}. The main drawback of classification models is that they require more training data. Xiao \etal~\cite{xiao2016learning} propose to combine multiple datasets for training and improve feature learning by domain guided dropout. Zheng \etal~\cite{Zheng2016,Zheng_2017_CVPR} point out that classification models are able to reach higher accuracy than siamese model, even without careful sample choosing. 

Recently, attention mechanism~\cite{Hydra,Liu2017,li2018harmonious,li2017learning,su2017pose} has been adopted to learn better features for person re-ID. For instance, HydraPlus-Net~\cite{Hydra} aggregates multiple feature layers within the spatial attentive regions extracted from multiple layers in the network. PDC model~\cite{su2017pose} enriches the person representation with pose-normalized images and re-weights the features by channel-wise attention. 
In this paper, we also formulate person re-identification as a classification problem, and we propose to emphasize foreground information in the aggregated representation by adding an axillary stream with spatial attention~(instance mask) and channel-wise re-weighting~(SEBlock), which is similar to HydraPlus-Net and PDC model. However, our work differs from them in that the attention mechanism in our work is introduced with a different motivation, which is to consider the foreground-background relationship instead of local-global or part-whole relationship. In addition, the architecture of our model is more clear and concise, along with more practical training strategy without multi-staged fine-tuning.

\section{Method}\label{sec:method}
As shown in Fig.~\ref{fig:pipeline}, our proposed person search method consists of two stages: pedestrian detection and re-identification.
In this section, we first give an overview of our framework, and then describe more details for both stages individually. 

\subsection{Overview}
A panoramic image is first fed into a pedestrian detector, which outputs several bounding boxes along with their confidence scores. We remove the bounding boxes whose confidence scores are lower than a given threshold. Only the remaining ones are used by the re-ID network. 

A post-processing is implemented on the detected persons before they are sent to the re-ID stage. Specifically, we expand each RoI (Region of Interest) with a ratio of $\gamma(\gamma >1)$ to include more context and crop out the person from the whole image. 
In order to separate the foreground person from background, we apply an off-the-shelf instance segmentation method FCIS~\cite{Li_2017_CVPR} on the whole image, and then designate the person to the right mask via majority vote. After that, for each person, we obtain a pair of images, one containing only the foreground person, and the other containing both the foreground and the background. See an illustration in Fig.~\ref{fig:vis_alg_1}.

Next up, in the re-ID model, the pair images go through two different paths, namely \textit{F-Net} and \textit{O-Net}, for individual modeling. 
Feature maps from the two paths are then concatenated and re-weighted by an SEBlock~\cite{hu2017}. After channel re-weighting, we pool the two dimensional feature maps into feature vectors using Global Average Pooling~(GAP). Finally, the feature vectors are projected to an L2-normalized $d$-dimensional subspace as the final identity descriptor. 

The pedestrian detector and re-ID model are trained independently. In order to avoid the mistakes resulting from the detector, we use the ground truth annotations instead of detections to train the re-ID model. 

\subsection{Pedestrian Detection}
We use a Faster R-CNN~\cite{Ren2017} detector for pedestrian detection. The Faster R-CNN architecture is composed of a base network for feature extraction, a region proposal network~(RPN) for proposal generation and a classification network for final predictions.

In this paper, we use VGG16~\cite{Simonyan2015} as our base network. The top convolutional layer `conv5\_3' produces 512 channels of feature maps, where the image resolution is reduced by a factor of 16. According to \cite{CityPersons2017cvpr}, up-sampling the input image is a reasonable way for compensation.

RPN is built upon `conv5\_3' to predict pedestrian candidate boxes. We follow the anchor settings in \cite{Ren2017} and set uniform scales ranging from the smallest and biggest persons we want to detect.
The RPN produces a large number of proposals, so we apply a humble Non-Maximum Suppression (NMS) with an intersection over union (IoU) threshold of 0.7 to remove repeating ones and also cut off low-scoring ones by a given threshold.

The remaining proposals are then sent to the classification network, where an RoI pooling layer ($512\times 7 \times 7$) is used to generate the same length of features for each proposal. The final detection confidence and corresponding bounding box regression parameters are regressed by fully connected layers. After bounding box regression, another NMS with IoU threshold of 0.45 is applied and low-scoring detections are cut off.

The base net, RPN and classification network are jointly trained using Stochastic Gradient Descent~(SGD).

\subsection{Person Re-ID via A Mask-guided Two-Stream CNN Model}
After RoIs for each person are obtained~(either from a detector or ground truth), we aim to extract discriminative features. First of all, we expand each RoI by a ratio of $\gamma$ to include more context. Then, we propose a two-stream structure to extract features for foreground person and whole image individually. The features from both streams are concatenated as enriched representations for the RoIs and a re-weighting operation is applied to highlight more informative features while suppressing less useful ones.

\begin{figure*}[t]
\vskip -20pt
\centering
\begin{minipage}{.5\linewidth}
\begin{algorithm}[H]
\caption{Foreground Separation}
\label{Alg}
\begin{flushleft}
        \textbf{Input:} RoI \textbf{b} $\in \mathbb{N}^4$, expand ratio $\gamma \in \mathbb{R}$, image \textbf{I} $\in \mathbb{R}^{h\times w \times 3}$, instance mask \textbf{M} $\in \mathbb{N}^{h\times w \time 3}$ \\
        \textbf{Output:} Masked image for an instance \textbf{I$^\prime_{\textbf{M}}$} $\in \mathbb{R}^{h^\prime \times w^\prime \times 3}$
\end{flushleft}
\begin{algorithmic}[1]
\STATE $\textbf{b}^\gamma \gets$ expand \textbf{b} by $\gamma$
\STATE Crop out image patch \textbf{I}$^\prime$ from image \textbf{I} according to $\textbf{b}^\gamma$
\STATE Crop out mask patch \textbf{M}$^\prime$ from mask \textbf{M} according to $\textbf{b}^\gamma$
\STATE Find the dominant instance $k$ inside \textbf{M}$^\prime$ by majority vote
\STATE Binarize \textbf{M}$^\prime$ by \textbf{M}$^\prime_b$ $\gets$ (\textbf{M}$^\prime == k$ )
\STATE Cast mask onto image by element-wise production:\\ \textbf{I$^\prime_{\textbf{M}}$} $\gets$ (\textbf{M}$^\prime_b \odot$ \textbf{I}$^\prime$)
\end{algorithmic}
\end{algorithm}
\end{minipage}\quad
\begin{minipage}{.45\linewidth}
\centering
\begin{figure}[H]
	  \centering
      \includegraphics[width=0.9\textwidth]{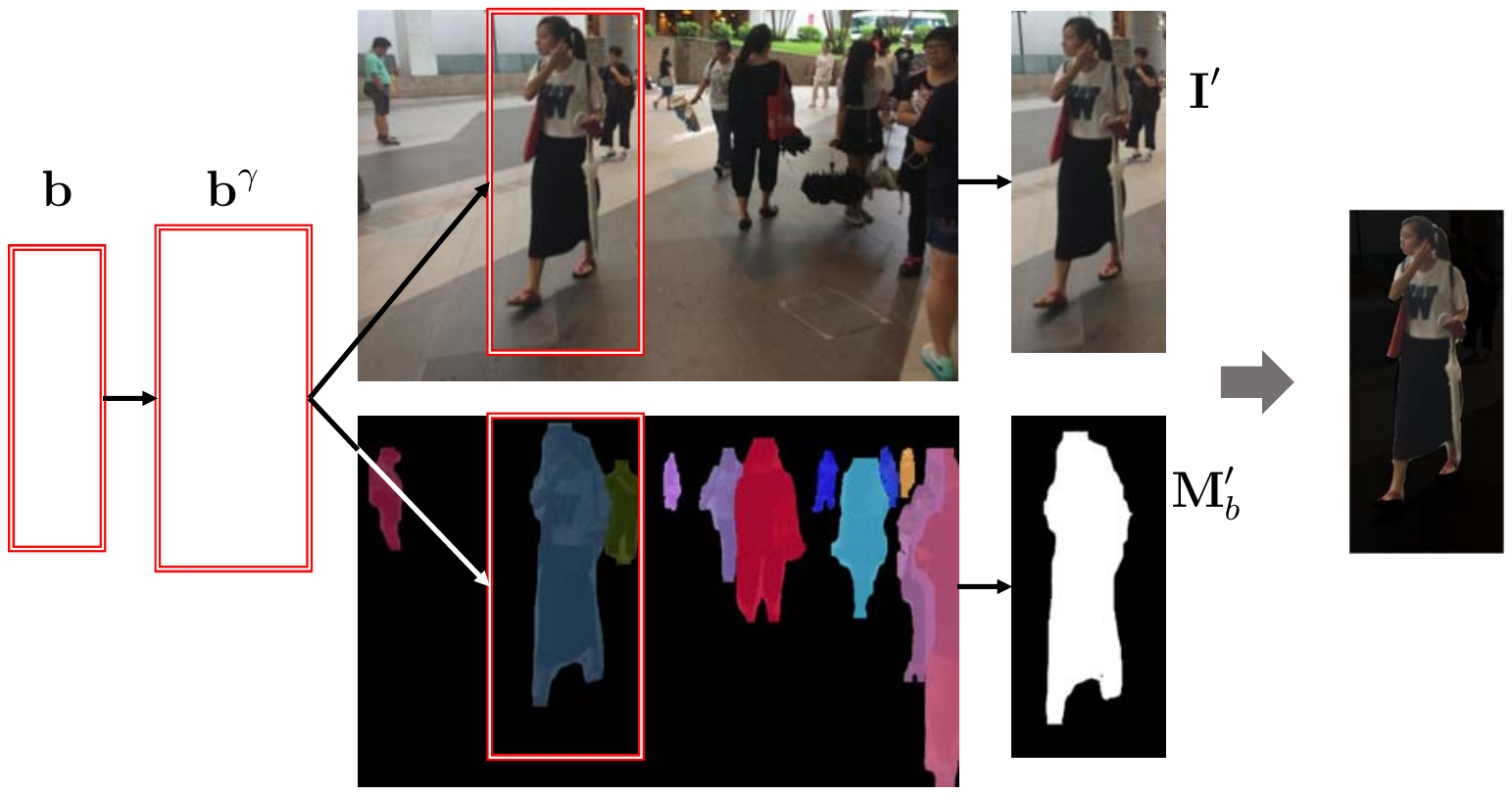}
      \caption{Illustration of foreground separation}
	  \label{fig:vis_alg_1}
\end{figure}
\begin{figure}[H]
	  \centering
      \includegraphics[width=0.9\textwidth]{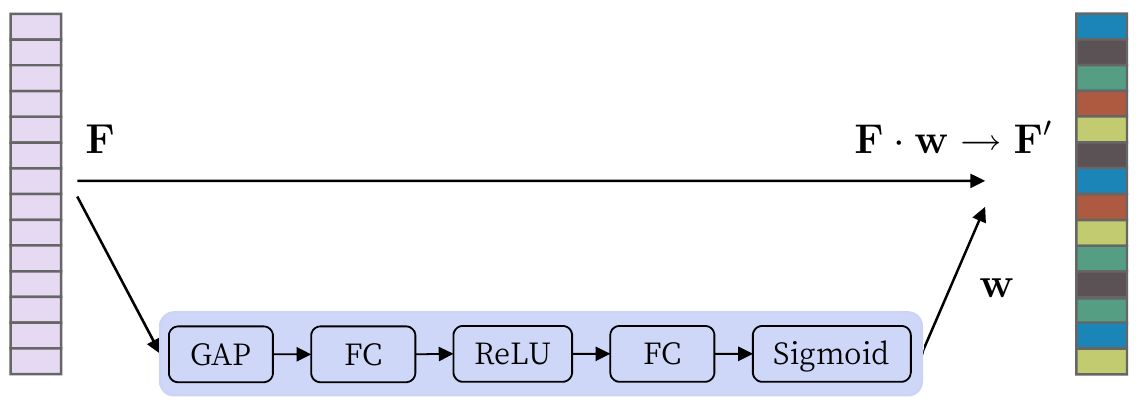} 
      \caption{Demonstration of an SEBlock~\cite{hu2017}}
      \label{fig:SEBlock}
\end{figure}

\end{minipage}

\end{figure*}

\noindent\textbf{Foreground separation.} The key step is to separate foreground and background for each RoI. We first apply an instance segmentation method FCIS~\cite{Li_2017_CVPR} on the whole image to obtain segmentation masks for persons. After that, we associate each RoI with its corresponding mask by majority vote. Those pixels inside and outside the mask boundary are considered as foreground and background respectively. 
We describe the detailed separation procedure in Algorithm \ref{Alg} and show an example in Fig.~\ref{fig:vis_alg_1}.

\noindent\textbf{Two-stream modeling.} After foreground separation, image pairs of each person are fed into the MGTS model. Specifically, foreground images go through F-Net and original images go through O-Net. Both F-Net and O-Net share the same architecture, but their network parameters are not shared. The corresponding feature maps, denoted as $\textbf{F}_{\textbf{F}}, \textbf{F}_{\textbf{O}} \in \mathbb{R}^{c \times h^{\prime\prime} \times w^{\prime\prime}}$ , are produced individually. Here $c$ denotes the number of channels, $h^{\prime\prime}, w^{\prime\prime}$ are the height and width of $\textbf{F}_{\textbf{F}}$ and $\textbf{F}_{\textbf{O}}$. The feature maps are then concatenated along the channel axis as $\textbf{F} \in \mathbb{R}^{ 2c \times h^{\prime\prime}\times w^{\prime\prime}}$. 

\noindent\textbf{Feature re-weighting.} 
We further re-weight all the feature maps with an SEBlock~\cite{hu2017}, which models the interdependencies between channels of convolutional features.
The architecture of an SEBlock is illustrated in Fig.~\ref{fig:SEBlock}. It is defined as a transformation from $\textbf{F}$ to $ \textbf{F}^\prime$:
\begin{align}
\textbf{F}^\prime &= \textbf{F} \cdot \textbf{w}, \quad \text{where} \\ \nonumber
\textbf{w} &=  [w_1, w_2, \hdots, w_{2c}], \quad w_i \in [0,1] \\ \nonumber
           &=\sigma(\textbf{W}_2\delta(\textbf{W}_1 f_{GAP}(\textbf{F}))),
\end{align}
$\sigma$ and $\delta$ refer to the Sigmoid activation and the ReLU~\cite{Nair2010} function respectively. $\textbf{W}_1$ and $\textbf{W}_2$ are the weight matrix of two FC layers. $f_{GAP}$ is the operation of GAP and $\cdot$ denotes channel-wise multiplication. SEBlock learns to selectively emphasis informative features and suppress less useful ones by re-weighting channel features using the weighting vector $\textbf{w}$.
In this way, foreground and background information are fully explored and re-calibrated, and hence help to optimize the final feature descriptor for person re-identification.

The re-weighted feature maps $\textbf{F}^\prime$ are then pooled to feature vectors $\textbf{f} \in \mathbb{R}^{2c}$ by GAP, and further projected to a L2-normalized $d$-dimensional subspace by an FC layer: 
\begin{align}
\textbf{x}^T = \dfrac{\textbf{W}^T\textbf{f}^T}{\|\textbf{W}^T\textbf{f}^T\|_2}, \textbf{W} \in \mathbb{R}^{2c\times d}.
\end{align}
The whole MGTS model is trained with ground truth RoIs and supervised by an Online Instance Matching loss~(OIM) \cite{Xiao_2017_CVPR}. The OIM objective is to maximize the expected log-likelihood:
\begin{align}
\mathcal{L} &= \mathbb{E}_\textbf{x}[\log p_t], \quad \text{where} \\ \nonumber
p_t &= \dfrac{\exp(\textbf{v}^T_t \textbf{x} / \tau)}{\sum_{j=1}^L\exp(\textbf{v}_j^T \textbf{x}/\tau) + \sum_{k=1}^Q\exp(\textbf{u}_k^T \textbf{x}/\tau)},
\end{align}
$p_t$ denotes the probability of $\textbf{x}$ belonging to class $t$. $\tau$ is a temperature factor similar to the one in Softmax function. $\textbf{v}_t$ is the class central feature vector of the $t$-th class. It is stored in a lookup table with size $L$ and incrementally updated during training with a momentum of $\eta$:
\begin{align}
\textbf{v}_t \gets \eta\textbf{v}_t + (1-\eta)\textbf{x},
\end{align}
where $\textbf{u}_k$ is a feature vector for an unlabeled person. A circular queue of size $Q$ is used to store $\textbf{u}_k$ vectors. It pops out old features and pushes in new features during training.

\begin{figure*}[t]
\begin{center}
\includegraphics[width=0.9\linewidth]{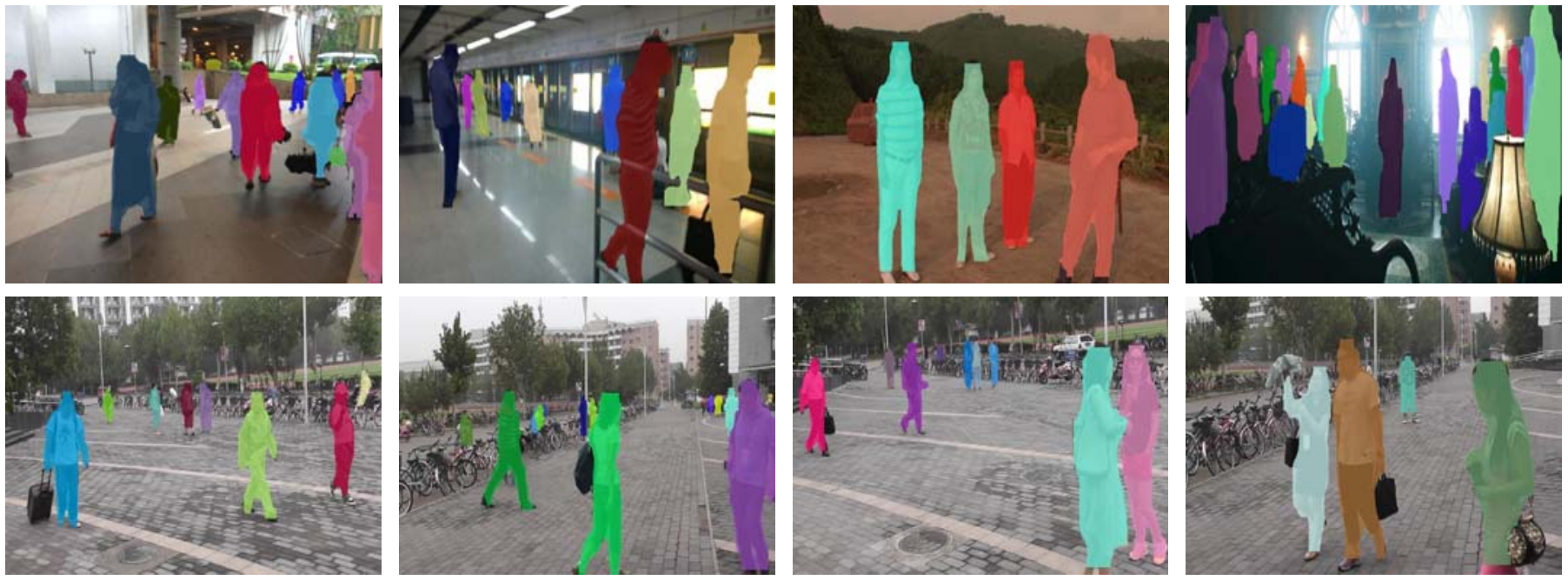}
\end{center}
   \caption{Instance segmentation results on CUHK-SYSU~(the first row) and PRW~(the second row) generated by the open sourced version of FCIS~\cite{Li_2017_CVPR}}
\label{fig:FCIS}
\end{figure*}

\section{Experiments}
In this section, we first introduce the datasets and evaluation protocols we use in our experiments, followed by some implementation details. After that, we show experimental results with comparison to state-of-the-art methods, followed by an ablation study to verify the design of our approach.

\subsection{Datasets}
\noindent\textbf{CUHK-SYSU.} CUHK-SYSU~\cite{Xiao_2017_CVPR} is a large-scale person search database consisted of street/urban scene images captured by a hand-held camera or selected from movie snapshots. It contains $18,184$ images and $96,143$ pedestrian bounding boxes. There are a total of $8,432$ labeled identities, and the rest of the pedestrians are served as negative samples for identification. We adopt the standard train/test split provided by the dataset, where the training set includes $11,206$ images and $5,532$ identities, while the testing set contains $2,900$ probe persons and $6,978$ gallery images. Moreover, each probe person corresponds to several gallery subsets with different sizes, which are defined in the dataset.

\noindent\textbf{PRW.} The PRW dataset~\cite{Zheng_2017_CVPR} is extracted from video frames captured with six cameras in a university campus. There are $11,816$ frames annotated with $34,304$ bounding boxes. Among all the pedestrians, $932$ identities are tagged and the rest of them are marked as unknown persons similar to CUHK-SYSU. The training set includes $5,134$ images with $482$ different persons. The testing set contains $2,057$ probe persons and $6,112$ gallery images. Different from CUHK-SYSU, the whole gallery set serves as the search space for each probe person.

\subsection{Evaluation Protocol}
\noindent\textbf{Pedestrian detection.} Average Precision~(AP) and recall are used to measure the performance of pedestrian detection. A detection bounding box is considered as a true positive if and only if its overlap ratio with any ground truth annotation is above $0.5$.

\noindent\textbf{Person search.} We adopt the mean Average Precision~(mAP) and cumulative matching characteristics~(CMC top-K) as performance metrics for re-identification and person search. The mAP metric reflects the accuracy and matching rate of searching a probe person from gallery images. CMC top-K is widely used for person re-identification task, where a matching is counted if there is at least one of the top-K predicted bounding boxes overlapping with the ground truth with an intersection-over-union~(IoU) larger than or equal to a threshold. The threshold is set to 0.5 throughout the paper.

\subsection{Implementation Details}
We implement our system with Pytorch. The VGG-based pedestrian detector is initialized with an ImageNet-pretrained model. It is trained using SGD with a batch size of 1. Input images are resized to have at least $900$ pixels on the short side and at most $1,500$ pixels on the long side. The initial learning rate is 0.001, decayed by a factor of 0.1 at 60K and 80K iterations respectively and kept unchanged until the model converges at 100K iterations. The first two convolutional blocks~(`conv1' and `conv2') are frozen during training, while other layers are updated. 

The RoI expand ratio $\gamma$ is set to 1.3. Both F-Net and O-Net of our MGTS model are based on ResNet50~\cite{He2016} and truncated at the last convolutional layer~(`conv5\_3'). The input image patches are re-scaled to an arbitrary size of $256\times 128$ and the batch size is set to 128. The model is trained with an initial learning rate of 0.001, decayed to 0.0001 after 11 epochs and kept identical until early cutting after epoch 15. The temperature scalar $\tau$, circular queue size $Q$ and momentum $\eta$ in OIM loss are set to $1/30$, $5000$ and $0.5$ respectively. The feature dimension $d$ is set to 128 through out the paper if not specified.

As of foreground extraction, we use the off-the-shelf instance segmentation method FCIS trained on COCO trainval35k~\cite{lin2014microsoft} without any fine-tuning\setcounter{footnote}{0}\footnote{\url{https://github.com/msracver/FCIS}}. Sample results of instance masks from CUHK-SYSU and PRW are shown in Fig.~\ref{fig:FCIS}, where we can see FCIS generalizes well to both datasets.

\subsection{Comparison with State-of-the-Art Methods}\label{sec:results}
In this subsection, we report our person search results on the CUHK-SYSU and PRW datasets, with a comparison to several state-of-the-art methods, including OIM~\cite{Xiao_2017_CVPR}, IAN~\cite{Xiao2017IANTI}, NPSM~\cite{Liu_2017_ICCV} and IDE~\cite{Zheng_2017_CVPR}.
Other than the above joint methods, we also compare with some methods which also solve the person search problem in two steps of pedestrian detection and person re-identification, similar to our method. These methods use different pedestrian detectors (DPM~\cite{DPM}, ACF~\cite{Dollar2014}, CCF~\cite{Yang2015}, LDCF~\cite{NIPS2014_5419}), person descriptors (BoW~\cite{Zheng2015}, LOMO~\cite{Liao_2015_CVPR}, DSIFT~\cite{Zhao2013}) and distance metrics (KISSME~\cite{Kostinger2012}, XQDA~\cite{Liao_2015_CVPR}).

\begin{figure}[t]
\vskip -30pt
\begin{minipage}[t]{.45\linewidth}
\begin{table}[H]
  \centering
  \caption{Comparison of results on CUHK-SYSU with gallery size of 100}
    \begin{tabular}{l|cc}
    \hline
    \textbf{Method} & \multicolumn{1}{c}{\textbf{mAP($\mathbf{\%}$)}} & \multicolumn{1}{c}{\textbf{top-1($\mathbf{\%}$)}} \bigstrut\\
    \hline
    CNN + DSIFT + Euclidean & 34.5  & 39.4  \bigstrut[t]\\
    CNN + DSIFT + KISSME & 47.8  & 53.6  \\
    CNN + BoW + Cosine & 56.9  & 62.3  \\
    CNN + LOMO + XQDA & 68.9  & 74.1  \\
    CNN + IDNet & 68.6  & 74.8  \bigstrut[b]\\
    \hline
    OIM~\cite{Xiao_2017_CVPR}   & 75.5  & 78.7  \bigstrut[t]\\
    IAN~\cite{Xiao2017IANTI}   & 76.3  & 80.1  \\
    NPSM~\cite{Liu_2017_ICCV}  & 77.9  & 81.2  \bigstrut[b]\\
    \hline
    Ours(CNN$_v$ + IDNetOIM) & 75.8 & 79.5 \bigstrut\\
	\textbf{Ours(CNN$_v$ + MGTS)} & \textbf{83.0}  & \textbf{83.7}  \bigstrut\\
    \hline
    \end{tabular}%
  \label{tab:Comparison_CUHK_SYSU}%
\end{table}%
\end{minipage}\qquad\quad
\begin{minipage}[t]{.50\linewidth}
\vskip +8mm
\begin{figure}[H]
\centering
\includegraphics[width=0.99\linewidth]{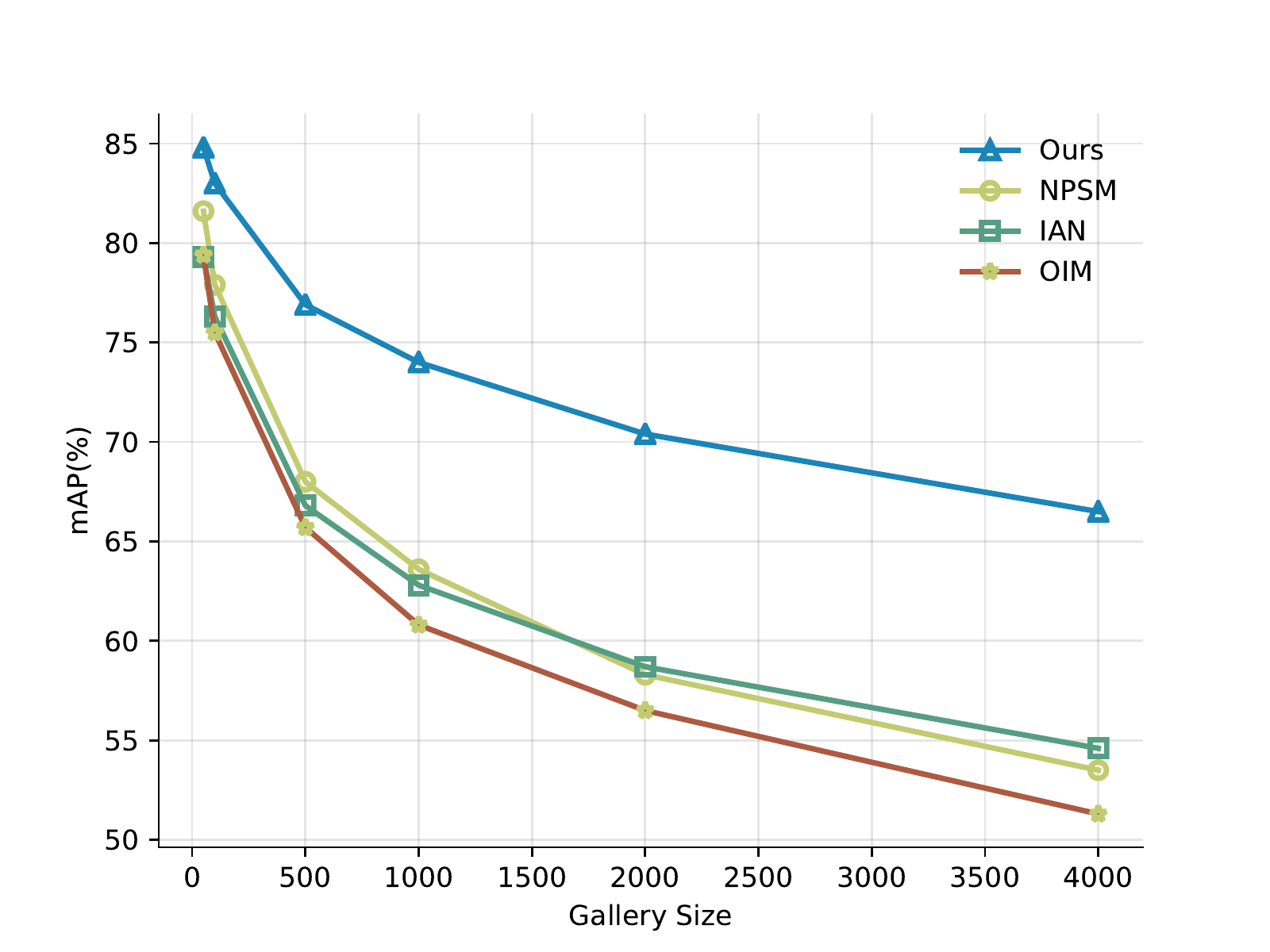}
\caption{Performance comparison on CUHK-SYSU with varying gallery sizes}
\label{fig:gallery_size}
\end{figure}
\end{minipage}
\vskip -10pt
\end{figure}

\noindent\textbf{Results on CUHK-SYSU.}\label{Results_on_CUHK_SYSU}
Table~\ref{tab:Comparison_CUHK_SYSU} shows the person search results on CUHK-SYSU with a gallery size of 100. We follow the notations defined in \cite{Xiao_2017_CVPR}, where ``CNN" denotes the Faster R-CNN detector based on ResNet50 and ``IDNet" represents a re-identification net defined in \cite{Xiao_2017_CVPR}. Our VGG-based detector is marked as ``CNN$_v$". IDNetOIM is a re-identification net trained with ground truth RoIs and supervised by an OIM loss. 
Compared with OIM, CNN$_v$ + IDNetOIM slightly improves the performance by solving detection and re-identification tasks in two independent models. By further employing our proposed MGTS model, we achieve $83.0\%$ mAP. 
Our final model outperforms the state-of-the-art method by more than $5$ pp w.r.t. mAP, and $2.5$ pp w.r.t. CMC top-1.

Moreover, we evaluate the proposed method~(CNN$_v$ + MGTS) under different gallery sizes along with other competitive methods. Figure \ref{fig:gallery_size} shows how the mAP changes with a varying gallery size of [50, 100, 500, 1000, 2000, 4000]. We can see that all methods suffer from a performance degeneration as the gallery size increases. However, our method outperforms others under different settings, which indicates the robustness of our method. Besides, we notice that the gap between our method and others become even larger as gallery size increases.  

\begin{figure}[t]
\centering
	\subfigure[]{
    \centering
	\includegraphics[height=4.2cm]{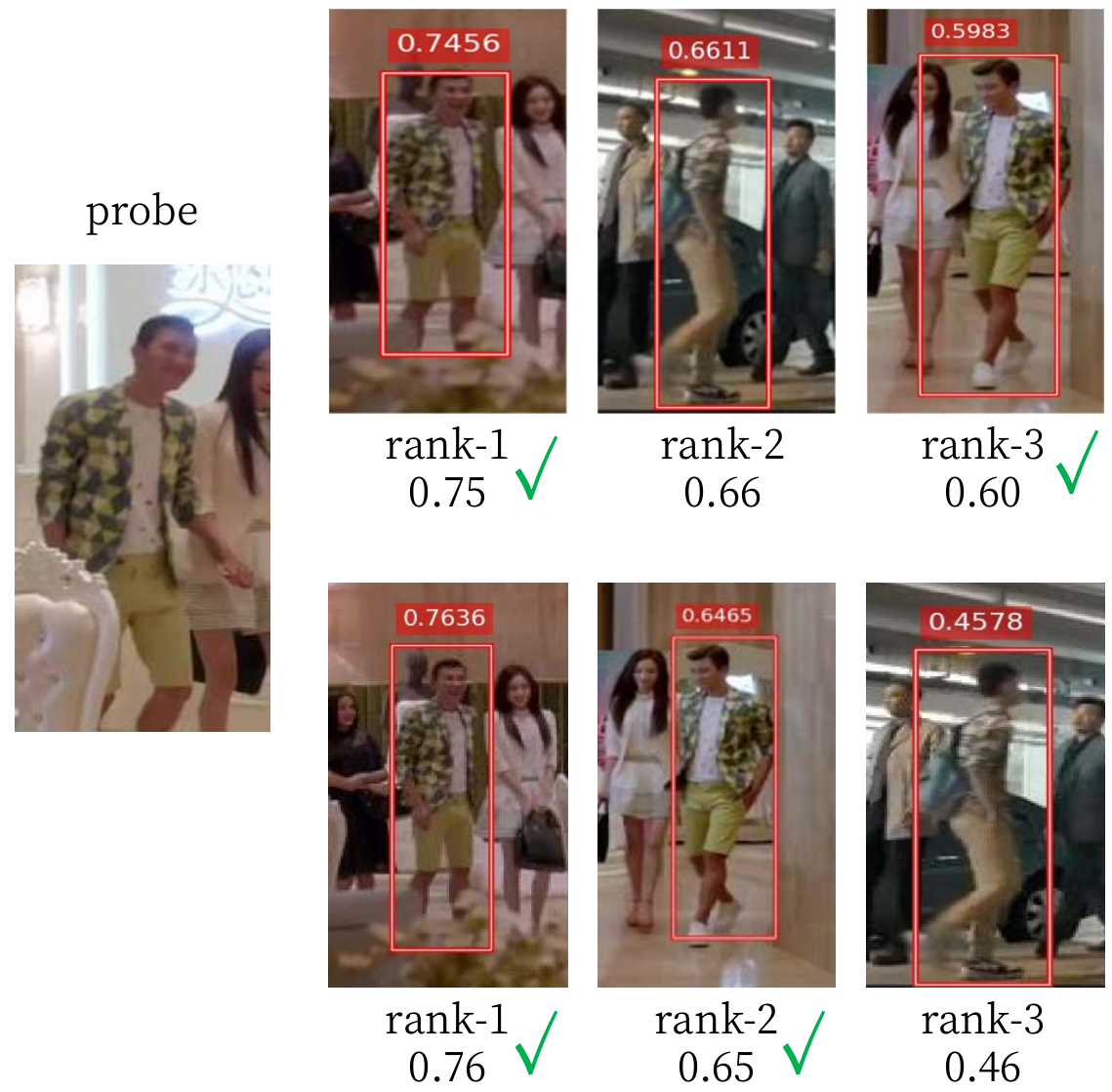}
    \label{case_1}
	}
    \quad
    \subfigure[]{
    \centering
	\includegraphics[height=4.2cm]{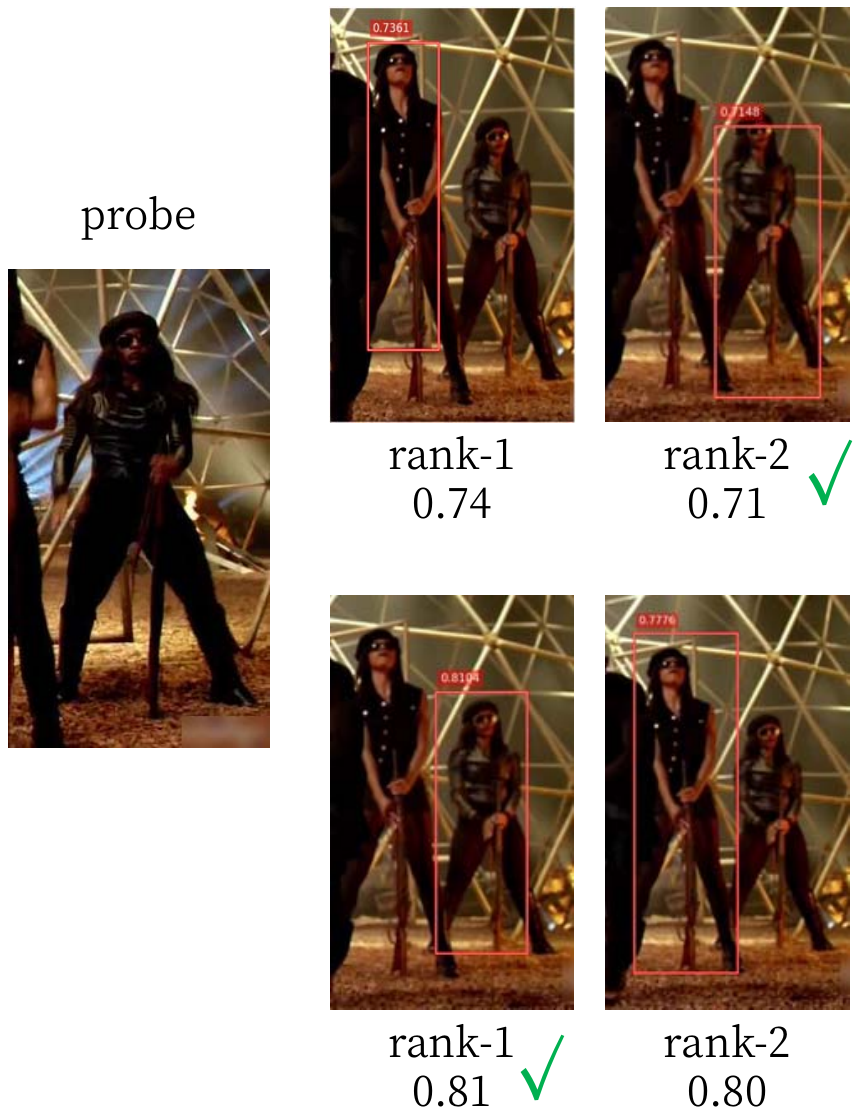}
    \label{case_2}
	}
    \quad
    \subfigure[]{
    \centering
	\includegraphics[height=4.2cm]{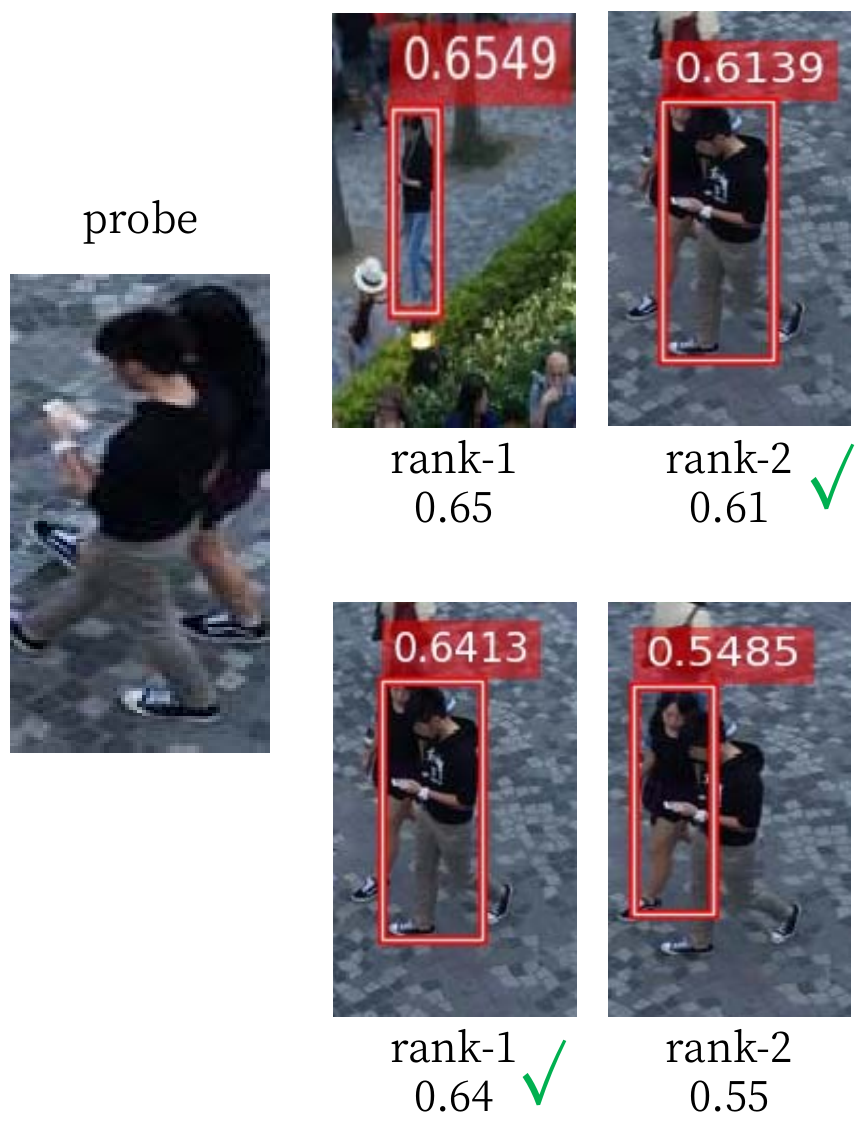}
    \label{case_5}
	}
    \quad
    \subfigure[]{
    \centering
	\includegraphics[height=4.2cm]{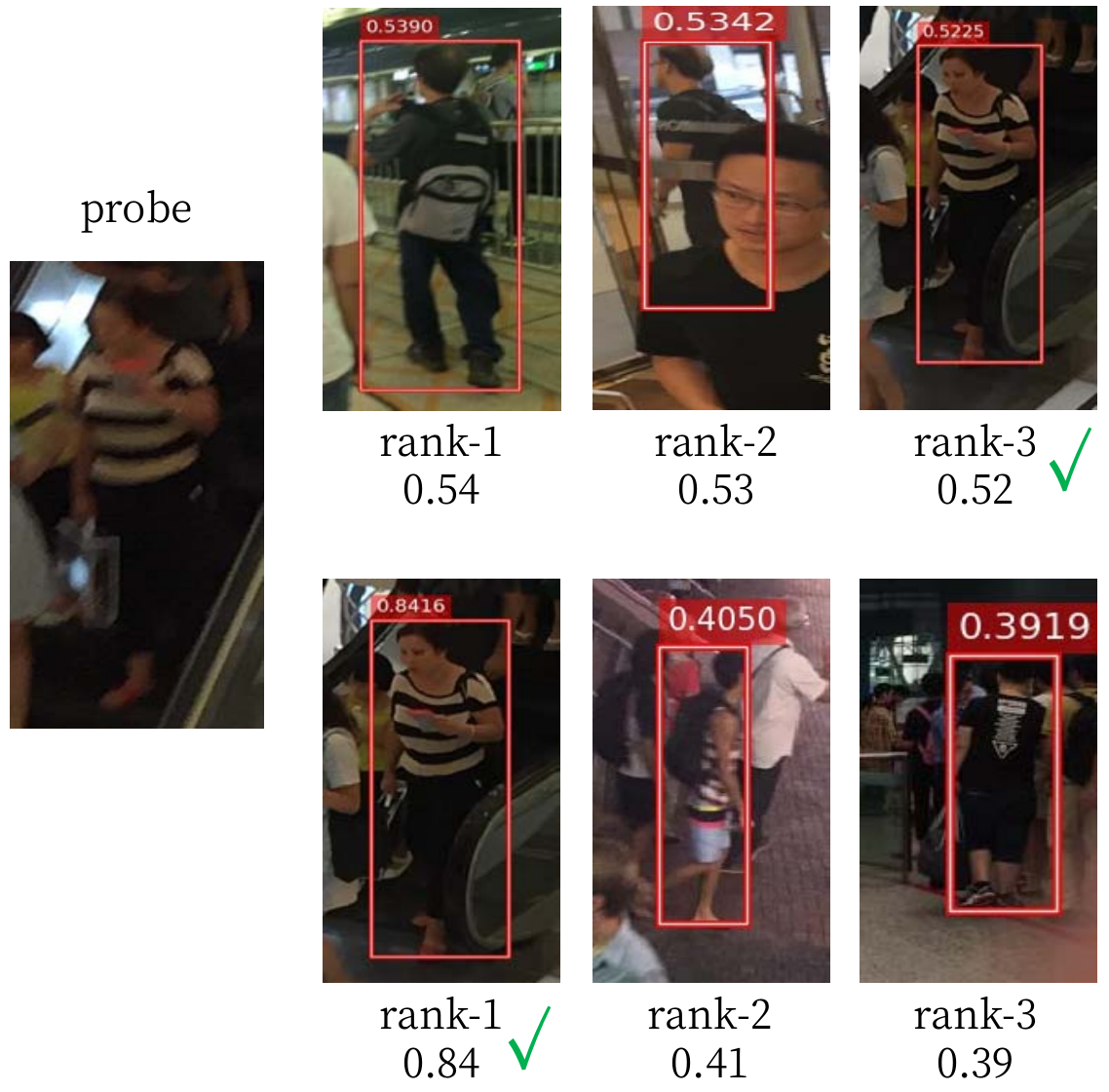}
    \label{case_3}
	}
    \quad
    \subfigure[]{
    \centering
	\includegraphics[height=4.2cm]{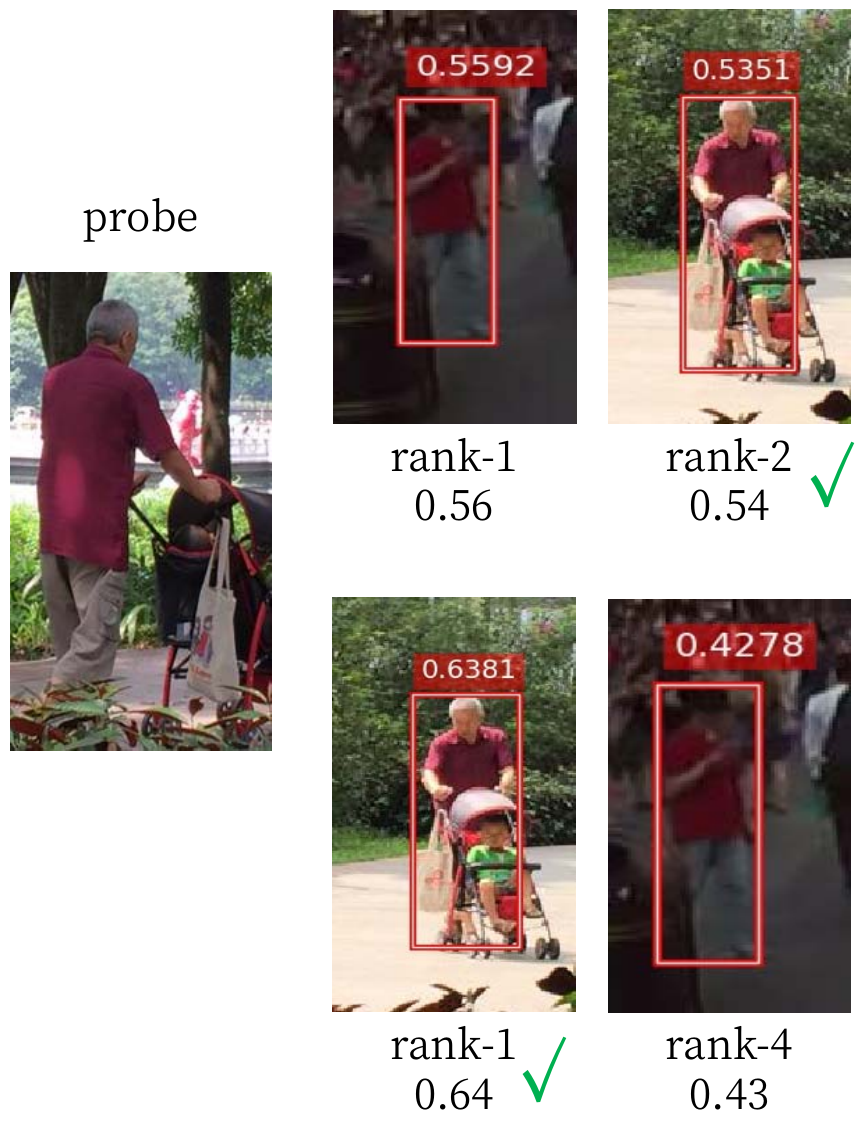}
    \label{case_4}}
    \quad
    \subfigure[]{
    \centering
	\includegraphics[height=4.2cm]{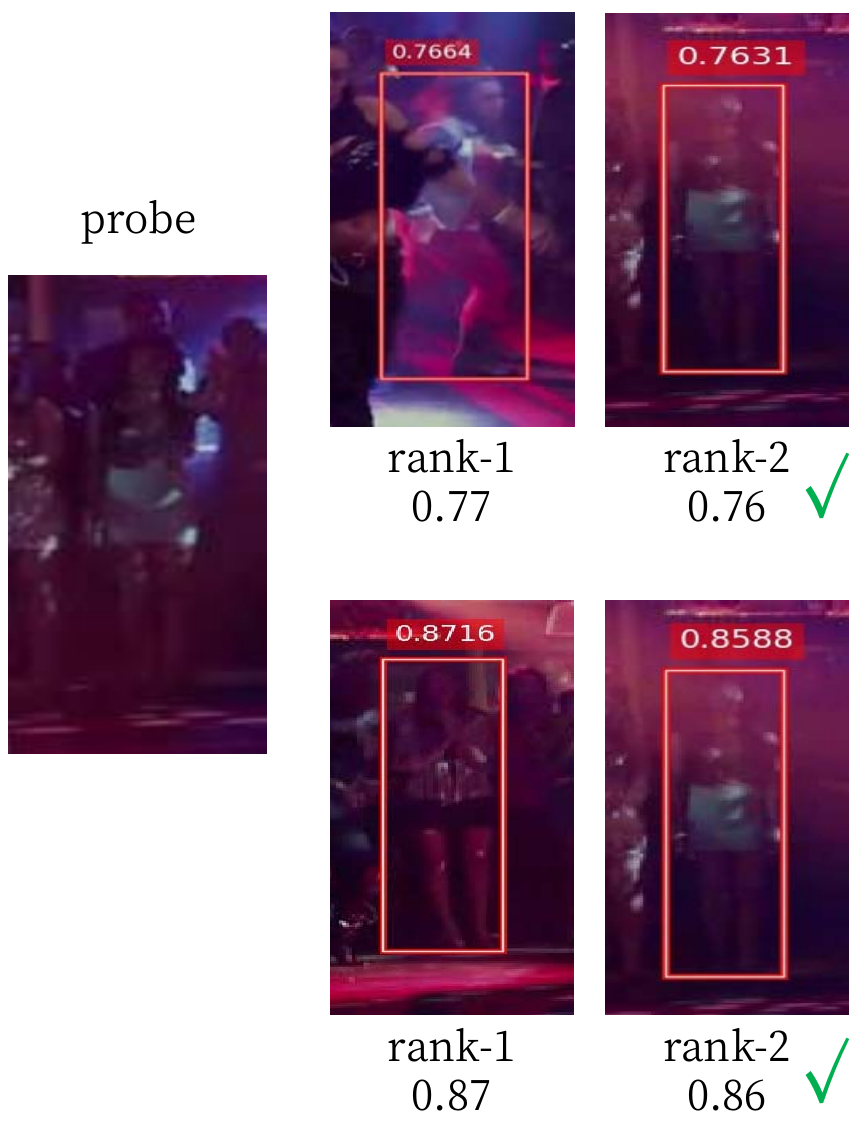}
    \label{case_6}
	}
\caption{Qualitative search results from OIM~\cite{Xiao_2017_CVPR} (upper row in each sub-figure) and our method (bottom row).  The ranking and similarity score are shown under each image patch; ``\cmark" denotes the correct matching. OIM would mistakenly assign high similarity to a wrong person wearing similar clothes to the probe, while our model successfully ranks it down. When faced with similar background, our method could emphasize the subtle difference on foreground people and return the correct ranking. (f) is a failure case with low visibility.}
\label{fig:case_study}
\end{figure}

We also show some qualitative results of our method and the competitive baseline OIM in Fig.~\ref{fig:case_study}. As can be seen in the figure, our method performs better on hard cases where gallery persons wear similar clothes with the probe person, possibly with the help of context information in the expanded RoI, \eg accompanied person~(Fig.~\ref{case_1}, \ref{case_3}), handrail~(Fig.~\ref{case_3}), baby carriage~(Fig.~\ref{case_4}) \etc. It is also more robust on hard cases where gallery entries share both similar foreground and background to the probe person~(Fig.\ref{case_2}, \ref{case_5}), where more subtle differences like hairstyle and gender shall be excavated from the emphasized foreground person. Fig.~\ref{case_6} shows a failure case where both OIM and MGTS suffer from bad illumination condition, which is rather challenging and needs more efforts in future works.

\begin{table}[t]
\vskip -10pt
  \centering
  \caption{Comparison of results on PRW}
    \begin{tabular}{l|cc}
    \hline
    \textbf{Method} & \multicolumn{1}{c}{\textbf{mAP($\mathbf{\%}$)}} & \multicolumn{1}{c}{\textbf{top-1($\mathbf{\%}$)}} \bigstrut\\
    \hline
    DPM-Alex + LOMO + XQDA & 13.0  & 34.1  \bigstrut[t]\\
    DPM-Alex + IDE$_{det}$ & 20.3  & 47.4  \\
    DPM-Alex + IDE$_{det}$ + CWS & 20.5  & 48.3  \bigstrut[b]\\
    \hline
    ACF-Alex + LOMO + XQDA & 10.3  & 30.6  \bigstrut[t]\\
    ACF-Alex + IDE$_{det}$ & 17.5  & 43.6  \\
    ACF-Alex + IDE$_{det}$ + CWS & 17.8  & 45.2  \bigstrut[b]\\
    \hline
    LDCF + LOMO + XQDA & 11.0  & 31.1  \bigstrut[t]\\
    LDCF + IDE$_{det}$ & 18.3  & 44.6  \\
    LDCF + IDE$_{det}$ + CWS & 18.3  & 45.5  \bigstrut[b]\\
    \hline
	OIM~\cite{Xiao_2017_CVPR}   & 21.3  & 49.9  \bigstrut[t]\\
	IAN~\cite{Xiao2017IANTI} & 23.0  & 61.9  \\
    NPSM~\cite{Liu_2017_ICCV}  & 24.2  & 53.1  \bigstrut[b]\\
    \hline
    Ours(CNN$_v$ + IDNetOIM) & 28.2 & 66.7 \bigstrut\\
	\textbf{Ours(CNN$_v$ + MGTS)} & \textbf{32.6}  & \textbf{72.1}  \bigstrut\\
    \hline
    \end{tabular}%
  \label{tab:Comparison_PRW}%
\end{table}%

\noindent\textbf{Results on PRW.}\label{Results_on_PRW}
In Table \ref{tab:Comparison_PRW} we report the evaluation results on the PRW dataset. A number of combinations of detection methods and re-identification models are explored in \cite{Zheng_2017_CVPR}. Among them, DPM~\cite{DPM} + AlexNet~\cite{AlexNet}-based R-CNN with ID-discriminative Embedding~(IDE$_{det}$) and Confidence Weighted Similarity(CWS) achieves the best performance.
In contrast, joint methods including OIM, IAN and NPSM, all achieve better results. But it is unclear whether the improvement comes from a joint solution or the usage of deeper networks (ResNet50/ResNet101) and a better performed detector (Faster R-CNN).

For fair comparison, we also employ ResNet50 and Faster R-CNN in our framework, and achieve significant improvements compared to joint methods. Specifically, we outperform the state of the art by $8.4$ pp  and $10.2$ pp  w.r.t. mAP and top-1 accuracy. These results again demonstrate the effectiveness of our proposed method.

\subsection{Ablation Study} 
From the experimental results in Sec.~\ref{sec:results}, we obtain significant improvement to our baseline method OIM~\cite{Xiao_2017_CVPR} on both standard benchmarks. The major differences between our method and OIM are as follows: (1) We solve the pedestrian detection and person re-identification tasks separately rather than jointly, i.e. we do not share features between them. (2) In the re-identification net, we model the foreground and original image in two parallel streams so as to obtain enriched representations.

In order to understand the impact of the above two changes, we conduct analytical experiments on CUHK-SYSU at a gallery size of 100, and provide discussions in the following.

\noindent\textbf{Integration vs. Separation.}
We investigate the impact of sharing features between detection and re-identification task on both performance.

In Table \ref{tab:IS_ablation_study_on_detector}, we compare the detection performance of a jointly trained model and a vanilla detector. We can see that the jointly trained detector under-performs the vanilla one by $8.5$ pp w.r.t. AP while reaching the same recall.

Similarly, we make a comparison on re-ID performance between a jointly trained model and a vanilla re-ID net in Table~\ref{tab:IS_ablation_study_on_reid}. The person search performance of a jointly trained OIM method is $0.6$ pp and $1.2$ pp lower in mAP and top-1 accuracy than a vanilla re-ID net~(IDNetOIM).

From the above comparisons, we conclude that joint training harms both the detection and re-ID performance, thus it is a better solution to solve them separately.

\begin{table}[t]
\vskip -10pt
\centering
\caption{Integration/Separation study on CUHK-SYSU with gallery size of 100. (a): Detector performance comparison between a jointly trained model and a vanilla detector; OIM-ours is our re-implementation of OIM. (b): Re-ID performance comparison for an integrated person search model and a na\"ive re-ID model}
  \subtable[]{
  	\centering
      \begin{tabular}{l||c| cc}
      \hline
      Method  & Joint  & \textbf{AP($\mathbf{\%}$)} & \textbf{Recall($\mathbf{\%}$)} \bigstrut\\
      \hline
      OIM-ours & \cmark & \multicolumn{1}{c}{69.5 }& \multicolumn{1}{c}{75.6 }  \bigstrut[t]\\
      CNN~\cite{Xiao_2017_CVPR}  &\xmark  & \multicolumn{1}{c}{78.0 } & \multicolumn{1}{c}{75.7 } \bigstrut[b]\\
      \hline
      \end{tabular}%
    \label{tab:IS_ablation_study_on_detector}%
  }  \qquad
  \subtable[]{
    \centering
      \begin{tabular}{l||c|cc}
      \hline
      Method & Joint & \textbf{mAP($\mathbf{\%}$)}   & \textbf{top-1($\mathbf{\%}$)} \bigstrut\\
      \hline
      GT + OIM~\cite{Xiao_2017_CVPR} & \cmark  & 77.9  & 80.5 \bigstrut[t]\\
      GT + IDNetOIM & \xmark & 78.5  & 81.7  \bigstrut[b]\\
      \hline
      \end{tabular}%
    \label{tab:IS_ablation_study_on_reid}%
  }
  \label{tab:IS_ablation_study}%
 \vskip -10pt
\end{table}

\noindent\textbf{Visual Component Study.}\label{fb_study}
In this part, we study the contributions of foreground and original image information to a re-ID system. To exclude the influence of detectors, all the following models are trained and tested using ground truth RoIs on CUHK-SYSU with a gallery size of 100. They are based on ResNet50 and supervised by an OIM loss. 

Four variants to the input RoI patch and their combinations are considered:
(1) Original RoI (\textbf{O});
(2) Masked foreground (\textbf{F});
(3) Masked background (\textbf{B});
(4) Expanded RoI (\textbf{E}) with a ratio of $\gamma$.

\begin{figure*}[t]
\vskip -10pt
\begin{minipage}[t]{.40\linewidth}
\vskip -15pt
\begin{table}[H]
  \centering
  \caption{Visual component study. Legend: \textbf{O}: Original image; \textbf{F}: Masked image with foreground people only; \textbf{B}: Masked image with background only; \textbf{E}: Expand RoI with a ratio of $\gamma$}
    \begin{tabular}{C{0.4cm}C{0.4cm}C{0.4cm}C{0.4cm}|cc}
    \hline
    \textbf{O} & \textbf{F} & \textbf{B} & \textbf{E} & \textbf{mAP($\mathbf{\%}$)} & \textbf{top-1($\mathbf{\%}$)} \bigstrut\\
    \hline
    \multicolumn{1}{c}{\cmark} & \multicolumn{1}{c}{} & \multicolumn{1}{c}{} & \multicolumn{1}{c|}{} & 78.5 & 81.7 \bigstrut[t]\\
    \multicolumn{1}{c}{} & \multicolumn{1}{c}{\cmark} & \multicolumn{1}{c}{} & \multicolumn{1}{c|}{} & 75.3  & 78.7 \\
    \multicolumn{1}{c}{} & \multicolumn{1}{c}{} & \multicolumn{1}{c}{\cmark} & \multicolumn{1}{c|}{} & 34.2  & 35.9 \\
    \multicolumn{1}{c}{\cmark} & \multicolumn{1}{c}{} & \multicolumn{1}{c}{} & \multicolumn{1}{c|}{\cmark} & 77.7 & 81.1 \\
    \multicolumn{1}{c}{} & \multicolumn{1}{c}{} & \multicolumn{1}{c}{\cmark} & \multicolumn{1}{c|}{\cmark} & 38.7 & 40.0 \\
    \hline
	\multicolumn{1}{c}{\cmark} & \multicolumn{1}{c}{\cmark} & \multicolumn{1}{c}{} & \multicolumn{1}{c|}{\cmark} & 89.1 & 90.0 \bigstrut[b]\\
    \hline
    \end{tabular}%
\label{tab:FB_ablation_study}%
\end{table}%
\end{minipage} \quad
\begin{minipage}[t]{.60\linewidth}
\begin{figure}[H]
\centering
\subfigure[]{
    \centering
	\includegraphics[height=2.5cm]{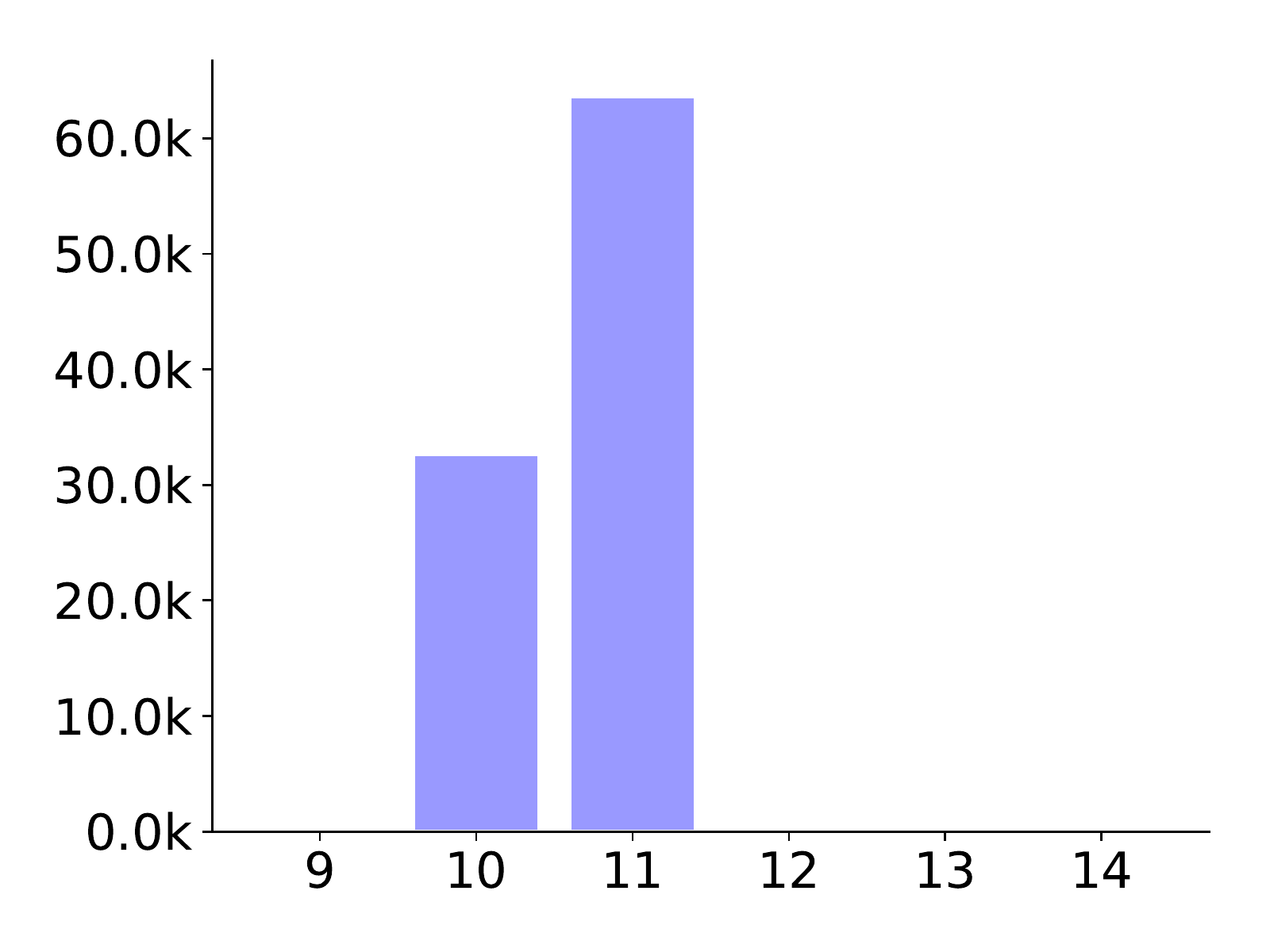}
	}
\subfigure[]{
    \centering
	\includegraphics[height=2.5cm]{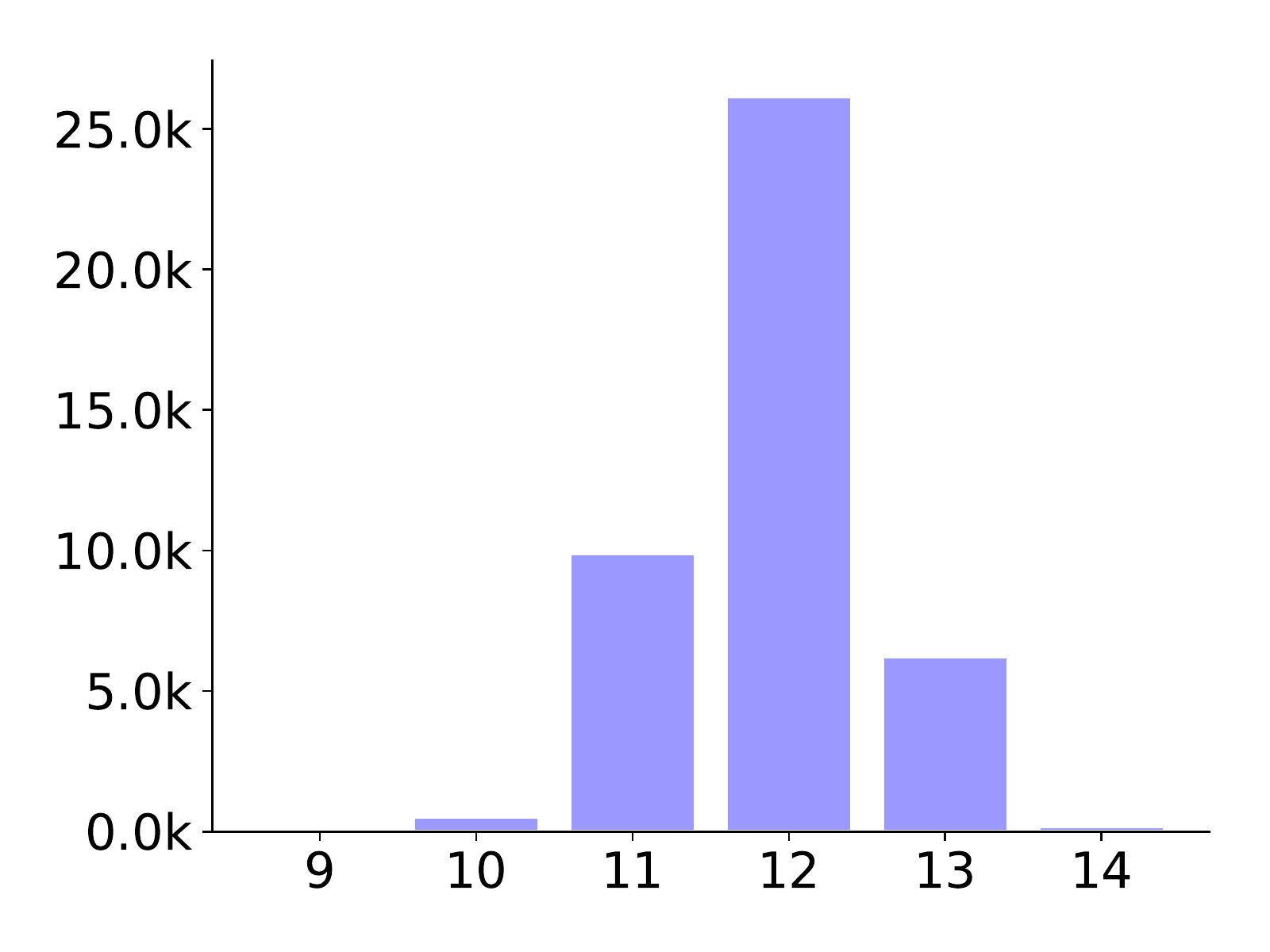}
	}
\caption{SEBlock weights statistics for CUHK-SYSU~(a) and PRW~(b). x-axis denotes $N^{20}(F)$: the number of foreground-related weights among the top 20 largest ones. y-axis denotes the number of training instances}
\label{fig:SE_statistics}
\end{figure}
\end{minipage}%
\vskip -10pt
\end{figure*}

Comparison results are shown in Table \ref{tab:FB_ablation_study}, from which we make the following observations:
\begin{enumerate}
\item Background is an important cue for re-ID. The mAP drops by $2.8$ pp when only the foreground is used, while discarding all background. More interestingly, using only background information yields an mAP of $34.2\%$, which can be further pushed to $38.7\%$ if RoI is expanded.
\item Modeling foreground and original image in two streams improves the results significantly. The two-stream model \textbf{O}+\textbf{F}+\textbf{E} reaches $89.1\%$ mAP, surpassing the one-stream model \textbf{O}+\textbf{E} by $11.4$ pp.
\end{enumerate}

\subsection{Model Inspection}
To further understand the respective impact of the two streams, we provide an analysis on the SEBlock weights of foreground vs. original image representations. The analysis is implemented based on the trained models in Sec.\ref{Results_on_CUHK_SYSU}, to which we feed all training samples in CUHK-SYSU~($96,143$ proposals) and PRW~($42,871$ proposals) respectively. 
For sample $i$, we compute three metrics: (1) average weight of F-Net channels, $Avg_i(F)$; (2) average weight of O-Net channels, $Avg_i(O)$; (3) number of channels that among top 20 of the whole network while come from F-Net, $N^{20}(F)$, the histograms of $N^{20}(F)$ for all training samples from two datasets are shown in Fig.~\ref{fig:SE_statistics}.

Based on analyzing the above statistics, we have the following findings:
\begin{enumerate}
\item The inequation $Avg_i(F) > Avg_i(O)$ holds for all samples. It demonstrates that in general the foreground patch contributes more than the original patch to the final feature vector, as it involves more informative cues for each identity.
\item From Fig.~\ref{fig:SE_statistics}, the majority of the top 20 channels come from the foreground stream for most samples. This observation indicates that the most informative cues are from the foreground patch.
\item Although the majority of the top 20 channels are represented by the foreground patch, we still observe that quite a few top channels are from the original patch. This is a good evidence showing the context information contained in the original image patch is helpful for the re-identification task.
\end{enumerate}

Moreover, the impact of the amount of context information is inspected by changing the RoI expansion value $\gamma$. We conduct a set of experiments on CUHK-SYSU with a gallery size of 100 and list the results in Table \ref{tab:roi_gammma}, from which we can draw the intuitive conclusion that \textbf{1)} the $\gamma$ is relatively stable when  $\gamma\in [1.2\  1.5]$; and \textbf{2)} a proper amount of context information is better than no context, while too much background could be harmful.

\begin{table}[t]
\vskip -10pt
\centering
\caption{Re-ID performance of MGTS on CUHK-SYSU with different $\gamma$}
\begin{tabular}{l|C{1.0cm}C{1.0cm}C{1.0cm}C{1.0cm}C{1.0cm}C{1.0cm}}
    \hline
    Value of $\gamma$ & 1.0   & 1.1   & 1.2   & 1.3   & 1.4   & 1.5  \bigstrut\\
    \hline
    \textbf{mAP~($\mathbf{\%}$)}   & 85.6  & 85.4  & 88.9  & \textbf{89.1}  & 87.8  & 87.1  \\
    \textbf{top-1~($\mathbf{\%}$)} & 86.6  & 86.2  & 89.8  & \textbf{90.0}  & 88.2  & 87.8  \\
    \hline
    \end{tabular}%
  \label{tab:roi_gammma}%
  \vskip -10pt
\end{table}

\subsection{Runtime Analysis and Acceleration}\label{sec:acceleration}
We implement our runtime analysis on a Tesla K80 GPU. For testing a $1500\times 900$ input image, our proposed approach takes around 1.3 seconds on average, including 626 ms for pedestrian detection, 579 ms for segmentation mask generation and another 64 ms for person re-identification. 

We notice half of the computational time is used to generate the segmentation mask. In order to accelerate, we propose an alternative option to use the tight ground truth bounding boxes as `weak' masks instead of the `accurate' FCIS masks.
The results are presented in Tab. \ref{tab:BoxMask}, from which we can see that using `accurate' FCIS masks yields better performance than using bounding box masks. However, using bounding boxes as weak masks still achieves promising results, which outperforms the single stream model without using any masks by a large margin ($\sim$ 7pp mAP) at comparable time cost. Therefore, our proposed method can be accelerated by a factor of $\sim$2x with an acceptable performance drop, while still surpassing the state-of-the-art results.

\begin{table}[t]
  \centering
  \caption{Comparison results on CUHK-SYSU with gallery size of 100. The three models are all trained and tested with ground truth bounding boxes. Overall runtime includes pedestrian detection, mask generation~(if used) and person feature extraction}
    \begin{tabular}{l|ccc}
    \hline    
    \textbf{Mask}  & \textbf{mAP~($\%$)}   & \textbf{top-1~($\%$)} \bigstrut & \textbf{Overall Runtime}~(s)\\
    \hline
    - & 78.5 & 81.7 &  0.65\\ \hline
    FCIS  & \textbf{89.1}  & \textbf{90.0} \bigstrut[t] & 1.27\\
    Bounding Box & 85.1  & 86.0 \bigstrut[b] & 0.69\\    
    \hline
    \end{tabular}%
\label{tab:BoxMask}%
\end{table}%

\section{Conclusion}
In this paper, we propose a novel deep learning based approach for person search. The task is accomplished in two steps: we first apply a Faster R-CNN detector for pedestrian detection on gallery images; and then make matchings between the probe image and output detections via a re-identification network. We obtain better results by training the detector and re-identification network separately, without sharing representations. We further enhance the re-identification accuracy by modeling the foreground and original image patches in two subnets to obtain enriched representations. 
Experimental results show that our proposed method significantly outperforms state-of-the-art methods on two standard benchmarks. 

Inspired by the success of utilizing the segmented foreground patch for additional feature extraction, for future work, we will explore to optimize the segmentation masks and identification accuracy in a joint framework, so as to obtain finer masks.

\section*{Acknowledgments}
This work was supported by the National Science Fund of China under Grant Nos. U1713208, 61472187 and 61702262, the 973 Program No.2014CB349303, Program for Changjiang Scholars, and ``the Fundamental Research Funds for the Central Universities'' No.30918011322.

\clearpage

\bibliographystyle{splncs}
\bibliography{egbib}

\begin{thebibliography}{10}

\bibitem{Xu2014}
Xu, Y., Ma, B., Huang, R., Lin, L.:
\newblock {Person search in a scene by jointly modeling people commonness and
  person uniqueness}.
\newblock In: ACM'MM. (2014)

\bibitem{Zheng_2017_CVPR}
Zheng, L., Zhang, H., Sun, S., Chandraker, M., Yang, Y., Tian, Q.:
\newblock Person re-identification in the wild.
\newblock In: CVPR. (2017)

\bibitem{Xiao_2017_CVPR}
Xiao, T., Li, S., Wang, B., Lin, L., Wang, X.:
\newblock Joint detection and identification feature learning for person
  search.
\newblock In: CVPR. (2017)

\bibitem{Xiao2017IANTI}
Xiao, J., Xie, Y., Tillo, T., Huang, K., Wei, Y., Feng, J.:
\newblock Ian: The individual aggregation network for person search.
\newblock arXiv preprint arXiv:1705.05552 (2017)

\bibitem{Liu_2017_ICCV}
Liu, H., Feng, J., Jie, Z., Jayashree, K., Zhao, B., Qi, M., Jiang, J., Yan,
  S.:
\newblock Neural person search machines.
\newblock In: ICCV. (2017)

\bibitem{Ren2017}
Ren, S., He, K., Girshick, R., Sun, J.:
\newblock {Faster R-CNN: Towards real-time object detection with region
  proposal networks}.
\newblock TPAMI \textbf{39}(6) (2017)

\bibitem{bg_removal_1}
Le, C.V., Hong, Q.N., Quang, T.T., Trung, N.D.:
\newblock Superpixel-based background removal for accuracy salience person
  re-identification.
\newblock In: ICCE-Asia. (2017)

\bibitem{bg_removal_2}
Nguyen, T.B., Pham, V.P., Le, T.L., Le, C.V.:
\newblock Background removal for improving saliency-based person
  re-identification.
\newblock In: KSE. (2016)

\bibitem{CityPersons2017cvpr}
Zhang, S., Benenson, R., Schiele, B.:
\newblock Citypersons: A diverse dataset for pedestrian detection.
\newblock In: CVPR. (2017)

\bibitem{Li_2017_CVPR}
Li, Y., Qi, H., Dai, J., Ji, X., Wei, Y.:
\newblock Fully convolutional instance-aware semantic segmentation.
\newblock In: CVPR. (2017)

\bibitem{Wen2016}
Wen, Y., Zhang, K., Li, Z., Qiao, Y.:
\newblock A discriminative feature learning approach for deep face recognition.
\newblock In: ECCV. (2016)

\bibitem{Dalal2005Cvpr}
Dalal, N., Triggs, B.:
\newblock Histograms of oriented gradients for human detection.
\newblock In: CVPR. (2005)

\bibitem{DPM}
Felzenszwalb, P.F., Girshick, R.B., McAllester, D., Ramanan, D.:
\newblock {Object detection with discriminatively trained part-based models}.
\newblock TPAMI (2010)

\bibitem{Dollar2009}
Dollar, P., Tu, Z., Perona, P., Belongie, S.:
\newblock {Integral channel features}.
\newblock In: BMVC. (2009)

\bibitem{shanshan2014cvpr}
Zhang, S., Bauckhage, C., Cremers, A.B.:
\newblock Informed haar-like features improve pedestrian detection.
\newblock In: CVPR. (2014)

\bibitem{ICF2015Cvpr}
Zhang, S., Benenson, R., Schiele, B.:
\newblock Filtered channel features for pedestrian detection.
\newblock In: CVPR. (2015)

\bibitem{shanshan_cvpr16}
Zhang, S., Benenson, R., Omran, M., Hosang, J., Schiele, B.:
\newblock How far are we from solving pedestrian detection?
\newblock In: CVPR. (2016)

\bibitem{shanshan2018pami}
Zhang, S., Benenson, R., Omran, M., Hosang, J., Schiele, B.:
\newblock Towards reaching human performance in pedestrian detection.
\newblock TPAMI (2018)

\bibitem{Ouyang2013JointDeep}
Ouyang, W., Wang, X.:
\newblock Joint deep learning for pedestrian detection.
\newblock In: ICCV. (2013)

\bibitem{Ouyang:DBNHuman}
Ouyang, W., Wang, X.:
\newblock A discriminative deep model for pedestrian detection with occlusion
  handling.
\newblock In: CVPR. (2012)

\bibitem{shanshan2018cvpr}
Zhang, S., Yang, J., Schiele, B.:
\newblock Occluded pedestrian detection through guided attention in cnns.
\newblock In: CVPR. (2018)

\bibitem{Wang2007}
Wang, X., Doretto, G., Sebastian, T., Rittscher, J., Tu, P.:
\newblock {Shape and appearance context modeling}.
\newblock In: ICCV. (2007)

\bibitem{conf/eccv/GrayT08}
Gray, D., Tao, H.:
\newblock Viewpoint invariant pedestrian recognition with an ensemble of
  localized features.
\newblock In: ECCV. (2008)

\bibitem{Farenzena2010}
Farenzena, M., Bazzani, L., Perina, A., Murino, V., Cristani, M.:
\newblock {Person re-identification by symmetry-driven accumulation of local
  features}.
\newblock In: CVPR. (2010)

\bibitem{Zhao2013}
Zhao, R., Ouyang, W., Wang, X.:
\newblock {Unsupervised salience learning for person re-identification}.
\newblock In: CVPR. (2013)

\bibitem{Liao_2015_CVPR}
Liao, S., Hu, Y., Zhu, X., Li, S.Z.:
\newblock Person re-identification by local maximal occurrence representation
  and metric learning.
\newblock In: CVPR. (2015)

\bibitem{zhao2017person}
Zhao, R., Oyang, W., Wang, X.:
\newblock Person re-identification by saliency learning.
\newblock IEEE transactions on pattern analysis and machine intelligence
  \textbf{39}(2) (2017)  356--370

\bibitem{Kostinger2012}
Kostinger, M., Hirzer, M., Wohlhart, P., Roth, P.M., Bischof, H.:
\newblock {Large scale metric learning from equivalence constraints}.
\newblock In: CVPR. (2012)

\bibitem{Li2015}
Li, X., Zheng, W.S., Wang, X., Xiang, T., Gong, S.:
\newblock {Multi-scale learning for low-resolution person re-identification}.
\newblock In: ICCV. (2015)

\bibitem{Liao2015}
Liao, S., Li, S.Z.:
\newblock {Efficient PSD constrained asymmetric metric learning for person
  re-identification}.
\newblock In: ICCV. (2015)

\bibitem{Paisitkriangkrai2015}
Paisitkriangkrai, S., Shen, C., {Van Den Hengel}, A.:
\newblock {Learning to rank in person re-identification with metric ensembles}.
\newblock In: CVPR. (2015)

\bibitem{Zhang_2016_CVPR}
Zhang, L., Xiang, T., Gong, S.:
\newblock Learning a discriminative null space for person re-identification.
\newblock In: CVPR. (2016)

\bibitem{Yi2014}
Yi, D., Lei, Z., Liao, S., Li, S.Z.:
\newblock {Deep metric learning for person re-identification}.
\newblock In: ICPR. (2014)

\bibitem{Li2014}
Li, W., Zhao, R., Xiao, T., Wang, X.:
\newblock {DeepReID: Deep filter pairing neural network for person
  re-identification}.
\newblock In: CVPR. (2014)

\bibitem{Ahmed2015}
Ahmed, E., Jones, M., Marks, T.K.:
\newblock {An improved deep learning architecture for person
  re-identification}.
\newblock In: CVPR. (2015)

\bibitem{Varior2016}
Varior, R.R., Shuai, B., Lu, J., Xu, D., Wang, G.:
\newblock {A siamese long short-term memory architecture for human
  re-identification}.
\newblock In: ECCV. (2016)

\bibitem{Liu2017}
Liu, H., Feng, J., Qi, M., Jiang, J., Yan, S.:
\newblock {End-to-end comparative attention networks for person
  re-identification}.
\newblock IEEE Transactions on Image Processing \textbf{26}(7) (2017)

\bibitem{xu2018attention}
Xu, J., Zhao, R., Zhu, F., Wang, H., Ouyang, W.:
\newblock Attention-aware compositional network for person re-identification.
\newblock In: CVPR. (2018)

\bibitem{Ding2015}
Ding, S., Lin, L., Wang, G., Chao, H.:
\newblock {Deep feature learning with relative distance comparison for person
  re-identification}.
\newblock Pattern Recognition \textbf{48}(10) (2015)

\bibitem{Cheng2016}
Cheng, D., Gong, Y., Zhou, S., Wang, J., Zheng, N.:
\newblock {Person re-identification by multi-channel parts-based CNN with
  improved triplet loss function}.
\newblock In: CVPR. (2016)

\bibitem{xiao2016learning}
Xiao, T., Li, H., Ouyang, W., Wang, X.:
\newblock Learning deep feature representations with domain guided dropout for
  person re-identification.
\newblock In: CVPR. (2016)

\bibitem{Zheng2016}
Zheng, L., Bie, Z., Sun, Y., Wang, J., Su, C., Wang, S., Tian, Q.:
\newblock {MARS}: A video benchmark for large-scale person re-identification.
\newblock In: ECCV. (2016)

\bibitem{Hydra}
Liu, X., Zhao, H., Tian, M., Sheng, L., Shao, J., Yi, S., Yan, J., Wang, X.:
\newblock Hydraplus-net: Attentive deep features for pedestrian analysis.
\newblock In: ICCV. (2017)

\bibitem{li2018harmonious}
Li, W., Zhu, X., Gong, S.:
\newblock Harmonious attention network for person re-identification.
\newblock arXiv preprint arXiv:1802.08122 (2018)

\bibitem{li2017learning}
Li, D., Chen, X., Zhang, Z., Huang, K.:
\newblock Learning deep context-aware features over body and latent parts for
  person re-identification.
\newblock In: CVPR. (2017)

\bibitem{su2017pose}
Su, C., Li, J., Zhang, S., Xing, J., Gao, W., Tian, Q.:
\newblock Pose-driven deep convolutional model for person re-identification.
\newblock In: ICCV. (2017)

\bibitem{hu2017}
Hu, J., Shen, L., Sun, G.:
\newblock Squeeze-and-excitation networks.
\newblock arXiv preprint arXiv:1709.01507 (2017)

\bibitem{Simonyan2015}
Simonyan, K., Zisserman, A.:
\newblock {Very deep convolutional networks for large-scale image recognition}.
\newblock In: ICLR. (2015)

\bibitem{Nair2010}
Nair, V., Hinton, G.E.:
\newblock {Rectified linear units improve restricted Boltzmann machines}.
\newblock In: ICML. (2010)

\bibitem{He2016}
He, K., Zhang, X., Ren, S., Sun, J.:
\newblock {Deep residual learning for image recognition}.
\newblock In: CVPR. (2016)

\bibitem{lin2014microsoft}
Lin, T.Y., Maire, M., Belongie, S., Hays, J., Perona, P., Ramanan, D.,
  Doll{\'a}r, P., Zitnick, C.L.:
\newblock Microsoft coco: Common objects in context.
\newblock In: ECCV. (2014)

\bibitem{Dollar2014}
Dollar, P., Appel, R., Belongie, S., Perona, P.:
\newblock {Fast feature pyramids for object detection}.
\newblock TPAMI \textbf{36}(8) (2014)

\bibitem{Yang2015}
Yang, B., Yan, J., Lei, Z., Li, S.Z.:
\newblock {Convolutional channel features}.
\newblock In: ICCV. (2015)

\bibitem{NIPS2014_5419}
Nam, W., Dollar, P., Han, J.H.:
\newblock Local decorrelation for improved pedestrian detection.
\newblock In: NIPS.
\newblock (2014)

\bibitem{Zheng2015}
Zheng, L., Shen, L., Tian, L., Wang, S., Wang, J., Tian, Q.:
\newblock {Scalable person re-identification: A benchmark}.
\newblock In: ICCV. (2015)

\bibitem{AlexNet}
Krizhevsky, A., Sutskever, I., Hinton, G.E.:
\newblock Imagenet classification with deep convolutional neural networks.
\newblock In: NIPS. (2012)

\end{thebibliography}
\end{document}